\pdfoutput=1

\documentclass[11pt]{article}

\usepackage{acl}

\usepackage{paperpkg}

\usepackage{times}
\usepackage{latexsym}
\usepackage{enumitem}
\usepackage{cleveref}
\usepackage{color, soul}
\crefname{section}{§}{§§}
\Crefname{section}{§}{§§}

\usepackage[T1]{fontenc}

\usepackage[utf8]{inputenc}

\usepackage{microtype}

\usepackage[colorinlistoftodos,prependcaption,textsize=tiny,disable]{todonotes}

\newcommand{\loraspecific}{{Kernel-wise}}
\newcommand{\loramix}{{Kernel-mix}}
\newcommand{\loraspecificlite}{{Kernel-wise-lite}}
\newcommand{\loramixlite}{{Kernel-mix-lite}}

\newcommand{\di}[1]{\todo[color=red!40] {Di: #1}}

\title{Empowering Parameter-Efficient Transfer Learning by Recognizing the Kernel Structure in Attention}

\author{Author 1 \and ... \and Author n \\
        Address line \\ ... \\ Address line}

\author{
Yifan Chen$^{1}$\thanks{~~Equal contribution. This work was performed while the first author was interning at Amazon Alexa AI.} \quad
{\bf Devamanyu Hazarika}$^{2}$\footnotemark[1] \quad
{\bf Mahdi Namazifar}$^{2}$ \\
{\bf Yang Liu}$^{2}$ \quad
{\bf Di Jin}$^{2}$\thanks{~~Correspondence to: Di Jin <djinamzn@amazon.com>} \quad
{\bf Dilek Hakkani-Tur}$^{2}$ \\
$^1$University of Illinois Urbana-Champaign \quad
$^2$Amazon Alexa AI
}

\begin{document}
\maketitle
\begin{abstract}
The massive amount of trainable parameters in the pre-trained language models (PLMs) makes them hard to be deployed to multiple downstream tasks. To address this issue, parameter-efficient transfer learning methods have been proposed to tune only a few parameters during fine-tuning while freezing the rest. This paper looks at existing methods along this line through the \textit{kernel lens}. Motivated by the connection between self-attention in transformer-based PLMs and kernel learning, we propose \textit{kernel-wise adapters}, namely \textit{Kernel-mix}, that utilize the kernel structure in self-attention to guide the assignment of the tunable parameters. These adapters use guidelines found in classical kernel learning and enable separate parameter tuning for each attention head. Our empirical results, over a diverse set of natural language generation and understanding tasks, show that our proposed adapters can attain or improve the strong performance of existing baselines.
\end{abstract}

\section{Introduction}
\label{sec:intro}

Transfer learning using large-scale transformer-based pre-trained language models (PLMs)~\citep{radford2019language} has become the standard scheme for various natural language processing (NLP) tasks. 
Among many strategies, fine-tuning these PLMs emerges as the predominant strategy to adapt the generic models to a specific task~\citep{howard2018universal}. 
However, deploying these models is a challenge as curating customized models across a wide variety of tasks would lead to scalability issues. 
It requires one to store (and sometimes move) multiple copies of the PLM parameters for different tasks, which is inefficient.

A popular approach to tackling such scalability issues is to make the PLM-based transfer learning more parameter-efficient. This can be done by freezing most of the PLM parameters and inserting small trainable modules into the PLM. Adapters~\citep{houlsby2019parameter, pfeiffer-etal-2020-mad,mahabadi2021compacter,hu2021lora} and  Prefix-/Prompt-tuning~\citep{shin2020eliciting,Li2021PrefixTuningOC, Lester2021ThePO, liu2021gpt, DBLP:journals/corr/abs-2110-07602} have emerged as the prominent approaches under this paradigm. These methods are incredibly parameter-efficient and have comparable performance to full fine-tuning models on many common NLP tasks (mainly in Natural Language Understanding) by tuning only 0.1-3\% task-specific parameters of the original PLMs.

However, most of these studies take the PLMs as a black box, i.e., these methods are not customized to transformers. This raises whether parameter-efficient fine-tuning has fully utilized the transformer structure in PLMs. Therefore in this work, we propose \textit{kernel-wise adaptation}, 
which recognizes and utilizes the kernel structure within self-attention---the core component in a transformer. 
Specifically, we take inspiration from recent work that connects self-attention to kernel learning~\citep{choromanski2020rethinking, chen2021skyformer, tsai2019transformer} to treat the different attention heads in a transformer's attention sub-layer as separate kernel estimators.
We hypothesize that parameter-efficient tuning can benefit from some useful guidelines in classical kernel learning literature and incorporate them into our proposed methods. These include:
\begin{enumerate}[leftmargin=*]
\item By interpreting attention heads as kernel estimators, we design the adaptation to be head-specific;
\item We assign more budgets of tunable parameters to learn the \textit{value} components in the attention mechanism, which correspond to \textit{coefficients} in kernel methods.
\end{enumerate}

We discuss these guidelines in detail in~\Cref{sec:guidelines}. We also evaluate our hypotheses through rigorous empirical evaluation. First, we test the effectiveness of the two guidelines above by comparing the default LoRA~\citep{hu2021lora}---a state-of-the-art approach for efficient adaptation---and two of its variants that implement the two guidelines, respectively. Next, we evaluate our variant of kernel-wise adaptation on three Natural Language Generation (NLG) benchmarks and two Natural Language Understanding (NLU) tasks using the GPT-2 architecture. 
While parameter-efficient work has extensively covered NLU tasks, it is unknown how well the results transfer to NLG tasks. As language generation typically requires more expressive models, we put more emphasis on multiple NLG tasks that include data-to-text, free-form question answering (QA), and summarization.
The empirical results in~\Cref{sec:results} demonstrate that with the same parameter budgets, our proposed method can attain better generation quality and classification accuracy than previous techniques,
and in many settings, it is close to or even outperforms the full parameter fine-tuning.

\section{Related Work}
\label{sec:related}

The literature on parameter-efficient adaptation can be broadly categorized as follows:

\textbf{Adapters}.
Originally proposed by~\citet{houlsby2019parameter, pfeiffer-etal-2020-mad}, adapters modulate the output of a transformer layer by inserting small Multi-layer Perception (MLP) bottleneck layers. Recent work has proposed many variants of the original adapters, including dropping adapters across several layers \citep{ruckle2020adapterdrop} or constraining adapters to be low-rank operators \citep{mahabadi2021compacter}.

A recent line of work focused on identifying the important subset of parameters within the PLMs.
\citet{ben2021bitfit} proposed to only tune the bias terms in the PLMs.
MPOP~\citep{liu2021enabling} suggested decomposing the weight matrices in PLMs through matrix product operators (MPO) and only trained the matrices of small size (freezing the large matrices) obtained from the decomposition,
which implicitly recognizes the small matrices as the important subset.

Low-rank adaptation (LoRA)~\citep{hu2021lora} directly assumes that the update of the weight matrices during training can be approximate low-rank,
and accordingly proposed to re-parameterize the original weight matrix $\mtx W$ by $\mtx W + \mtx{BA}$,
where $\mtx W$ is frozen to its pre-trained weights whereas $\mtx{A} \in \mb R^{r \times N_h p}, \mtx{B} \in \mb R^{N_h p \times r}$ are updated in training \footnote{In implementation, LoRA also trains the bias terms in the linear transform besides the matrices $\mtx{A}, \mtx{B}$, 
while for brevity, the bias terms are omitted throughout the paper.}. 
We note that LoRA too introduces new weights $\mtx{A}$ and $\mtx{B}$ similar to adapters, but they are used only to re-parameterize the existing weights and do not add extra sandwiched layers that modify the original model architecture.

\paragraph{Prefix-tuning.} Originally shown in GPT-3 \citep{brown2020language}, 
prompts are extra tokens that help in the task adaptation of PLMs. Transitioning from the manual design of prompts, \citet{shin2020eliciting} searched for the prompts over the discrete space of tokens based on the task-specific training data; \citet{Li2021PrefixTuningOC, Lester2021ThePO, liu2021gpt, DBLP:journals/corr/abs-2110-07602} further extended the search space to continuous prompts and tuned the prompts through back-propagating the error in training. Prompt-based methods have been shown to be similar to adapters by~\citep{he2021towards}. 

\section{Preliminaries}
\label{sec:prelims}

We start by providing a brief introduction to the transformer architecture (\Cref{sec:transformer}) and then revisit the connection between attention and kernel estimators (\Cref{sec:attn-as-kernel}), building on which we propose the kernel-wise adapter in \Cref{sec:method}.

\subsection{Transformer Architecture}
\label{sec:transformer}

Transformers~\citep{vaswani2017attention} are composed of $L$ stacked layers, where each layer comprises of a multi-headed attention and a fully connected feed-forward network (FFN) sub-layer.\footnote{For simplicity we omit the cross-attention module in 
transformer-based encoder-decoder models.} The attention sub-layer,
assuming $N_h$ heads and dimension size $p$ for each head, first maps an input $\mtx{X} \in \mb R^{n \times N_h p}$ into the query $(\mtx{Q})$, key $(\mtx{K})$, and value $(\mtx{V})$ matrices through the following affine transformations:
\begin{align}
\mtx{Q/K/V} &= \mtx{X} \mtx{W}_{[q/k/v]} + \mtx{1} \mtx{b}_{[q/k/v]}^T, \label{eqn:linear_transform}
\end{align}
where $\mtx{Q}, \mtx{K}, \mtx{V} \in \mb R^{n \times N_hp}$, $\mtx{W}_q, \mtx{W}_k, \mtx{W}_v$ are $N_h p$-by-$N_h p$ weight matrices, 
and $\mtx{b}_q, \mtx{b}_k, \mtx{b}_v \in \mb R^{N_h p}$ are the bias terms\footnote{To ease the notations we adopt the setting where $\mtx{X}, \mtx{Q}, \mtx{K}, \mtx{V}$ have the same shape.}.
After the transformation, the three components $\mtx{Q}, \mtx{K}, \mtx{V}$ are split into $N_h$ blocks corresponding to different heads. For example, $\mtx{Q}$ is re-written as $\mtx{Q} = \left( \mtx{Q}^{(1)}, \cdots, \mtx{Q}^{(N_h)} \right)$, where each block $\mtx{Q}^{(h)} = \mtx{X} \mtx{W}_q^{(h)} + \mtx{1} (\mtx{b}_q^{(h)})^T$ is an $n$-by-$p$ matrix,
and $\mtx{W}_q^{(h)}, \mtx{b}_q^{(h)}$ are the corresponding parts in $\mtx{W}_q, \mtx{b}_q$. The attention output for the $h^{th}$ head is then computed as:
\begin{align}
\mtx{L}^{(h)} \mtx{V}^{(h)} &\defeq \text{softmax}(\mtx{Q}^{(h)} (\mtx{K}^{(h)})^T / \sqrt{p}) \mtx{V}^{(h)} \nonumber \\ &= (\mtx{D}^{(h)})^{-1} \mtx{M}^{(h)} \mtx{V}^{(h)},
\label{eqn:attn}
\end{align}
where $\mtx{M}^{(h)} \defeq \exp\left(\mtx{Q}^{(h)} (\mtx{K}^{(h)})^T / \sqrt{p}\right)$ 
and $\mtx{D}^{(h)}$ is a diagonal matrix in which $\mtx{D}^{(h)}_{ii}$ is the sum of the $i$-th row in $\mtx{M}^{(h)}$, corresponding to the normalization part in softmax.

After we obtain the outputs in each head,
they are concatenated as,
\begin{equation*}
    \mtx{L} \defeq (\mtx{L}^{(1)} \mtx{V}^{(1)}, \dots, \mtx{L}^{(N_h)} \mtx{V}^{(N_h)}),
\end{equation*}
followed by the overall output,
\begin{align}
\mtx{L} \mtx{W}_o + \mtx{1} \mtx{b}_o^T,
\end{align}
where $\mtx{W}_o$ and $\mtx{b}_o$ are similarly sized as the other matrices in~\Cref{eqn:linear_transform}.

\subsection{Attention as Kernel Estimators}
\label{sec:attn-as-kernel}

For each head in the attention module, we have given the expression of attention output in Equation~\ref{eqn:attn}.
In this subsection, we will re-write attention as a kernel estimator to show the connection.

In computing the attention output (of a single head), we have a length-$n$ input sequence $\{x_i\}_{i=1}^n$ (the rows in $\mtx{X}$) and accordingly we can obtain $N$ \footnote{Note that $N$ may not always equal $n$, such as in cross attention ($N \neq n$) or in prefix-tuning ($N > n$ due to the prefix prepended to the key matrix) \citep{Li2021PrefixTuningOC}.} key vectors $\{k_j\}_{j=1}^N \subset \mb R^p$ (from the \textit{key} matrix $\mtx{K}$) and query vectors $\{q_i\}_{i=1}^n \subset \mb R^p$ (from $\mtx{Q}$).\footnote{In this subsection we omit the superscript $(h)$ for simplicity since the discussion is limited within a single head}
The original goal of self-attention is to obtain the representation of each input token $x_i$: $g(x_i)$.
By denotation exchange: $q_i:=x_i$ and $f(q_i):=g(x_i)$, we can also understand the aforementioned self-attention module as returning the representation $f(q_i)$ of the input query vector $q_i$ through $\{k_j\}_{j=1}^n$,
which behaves as a kernel estimator~\citep{choromanski2020rethinking, peng2020random, chen2021skyformer}.
Specifically, for a single query vector $q_i$, a Nadaraya–Watson kernel estimator \citep[Definition~5.39]{wasserman2006all} models its representation as,
\begin{align}
\label{eq:kernel_estimator}
& f(q_i) = \sum_{j=1}^n \ell_j(q_i) c_j, \\
& \text{where} \quad \ell_j(q_i) \defeq \frac{\kappa(q_i, k_j)}{\sum_{k=1}^N \kappa(q_i, k_j)}. \nonumber
\end{align}
Here, $\kappa(\cdot, \cdot)$ is a kernel function, and $c_j$'s are the coefficients ($c_j$ can either be a scalar or a vector in different applications) that are learned during training. In this estimator, $\{k_j\}_{j=1}^n$ serve as the \textit{supporting points} which help construct the representation for an input $q_i$.

For kernel function $\kappa(x, y) = \exp\left(\dotp{x}{y} / \sqrt{p} \right)$,
we slightly abuse the notation $\kappa(\mtx{Q}, \mtx{K})$ to represent an $n$-by-$N$ empirical kernel matrix, whose element in the $i$-th row and the $j$-th column is $\kappa(q_i, k_j), \forall i \in [n], j \in [N]$.
With these notations, the representation of the whole sequence $\mtx{Q}$ will be,
\begin{align}
\label{eq:matrix_form}
\mtx{D}^{-1} \kappa(\mtx{Q}, \mtx{K}) \mtx{C},
\end{align}
where $\mtx{D}$ is a diagonal matrix for row normalization in Eq.~(\ref{eq:kernel_estimator}), and $\mtx{C}$ is an $N$-by-$p$ matrix whose $j$-th row is $c_j$.
Considering the correspondence between Equation~(\ref{eq:matrix_form}) and the standard softmax attention in Equation~(\ref{eqn:attn}),
we can have a finer division of the attention module: the empirical kernel matrix $\kappa(\mtx{Q}, \mtx{K})$ ($\mtx{D}$ is decided by $\kappa(\mtx{Q}, \mtx{K})$) and the coefficient part $\mtx{C}$,
which includes but is \textit{not} limited to \textit{value} matrices in attention (see~\Cref{sec:method}). 
In what follows, we will discuss how we build adapters for these two parts.

\section{Method}
\label{sec:method}

We introduce our \textit{\loramix} method in this section, which builds upon the proposed \textit{\loraspecific} adaptation. 
To explain the principle behind \loraspecific, we first illustrate the guidelines we aim to adopt in our adapter design and show that existing methods fail to satisfy them (\Cref{sec:guidelines}). 
With these details, we finally propose our method in \Cref{subsec:method} and \Cref{sec:final-model}.

\subsection{\textit{Guidelines} Motivated by Kernel Learning}
\label{sec:guidelines}

Given the analogy between attention in PLMs and kernel estimators, we hypothesize that parameter-efficient adaptation should be aware of this connection in transformer-based PLMs and utilize desirable guidelines emerging from the literature on kernel learning. Here we discuss the guidelines introduced in~\Cref{sec:intro} in further detail. 

Guideline-1 suggests that the adaptation should be head-specific.
Conceptually, different heads correspond to different empirical kernel matrices (distinct distribution of attention scores),
and it will be beneficial to adapt the attention module in a head-specific manner.
The effect of head-specific adaptation is also observed by other work, e.g., \citep{he2021towards} that mentioned multi-head influence can make methods such as prefix-tuning more expressive.

Guideline-2 is that we should assign more parameter budgets to the coefficient (or \textit{value}) part of attention compared to the empirical kernel matrix part (\textit{query} and \textit{key}).
This guideline comes from the classical optimization procedure in kernel learning~\citep[Definition~5.29]{wasserman2006all} where we fix the kernel in use and only perform the unconstrained optimization for the coefficients ($c_j$'s in~\Cref{eq:kernel_estimator}).
This practice in kernel learning can be justified by Representer Theorem \citep{scholkopf2001generalized} that the minimizer $f^*$ of some certain empirical risks admits a representation of the form: 
\begin{align*}
    f^*(\cdot) = \sum_{j=1}^N \alpha_j \kappa(\cdot, k_j),
\end{align*}
where $\alpha_j$'s are the free parameters to optimize.
Representer Theorem indicates that the target estimator $f^*(\cdot)$ is simply a linear combination of $\kappa(\cdot, k_j)$'s, 
and therefore many kernel methods focus on optimizing the coefficients $\alpha_j$'s.
Revisiting \Cref{eq:kernel_estimator},
we find $\alpha_j = \frac{c_j}{\sum_{j'=1}^N \kappa(q_i, k_{j'})}$ under the setting of transformers,
which motivates us to apply Guideline-2 to better model the tunable coefficient part $c_j$'s in attention.

In addition, kernel learning theory concludes that the sample efficiency of a Nadaraya-Watson kernel estimator is mainly influenced by its bandwidth (the scaling factor, corresponding to the factor $\frac{1}{\sqrt{p}}$ in Equation~(\ref{eqn:attn})), 
rather than the concrete form of the empirical kernel matrix $\kappa(\mtx{Q}, \mtx{K})$~\citep[Section~5.4]{wasserman2006all}.
This implies that the adaptation to the empirical kernel matrix can be conservative. 
This is also similar to the conjecture by \cite{he2021towards}, which mentions ``attention learns pairwise positional interactions which do not require large capacity for adapting to new tasks.''

\paragraph{Do Existing Adapters Satisfy the Guidelines?}

Adapters are designed to modify the hidden states in a certain step in a transformer, and their mechanism can be stated as,
\begin{align*}
\mtx H \leftarrow \mtx H + \Delta \mtx H,
\end{align*}
where $\mtx H$ is the ``hidden state'' in a certain step, and $\Delta \mtx H$ is the update given by the adapter. 
As shown in \citep{he2021towards}, this definition embodies most of the recent proposals for efficient adaptation, such as adapters, prefix-tuning, LoRA, and similar variants.  

The original MLP-based adapters, which only adjust the output of a particular layer \citep{houlsby2019parameter}, do not modify the empirical kernel matrix in the attention sub-layer.

Prefix-tuning satisfies the first guideline as it is head-specific by nature (it prepends trainable continuous prefixes to both key and value matrices in each head). However, it fails to satisfy the second guideline as it enforces an equal assignment of tunable parameters to both the kernel and the coefficient parts (since the prefixes for \textit{key} matrices and \textit{value} matrices correspond to each other). 

As for weight-updating adapters, such as LoRA~\citep{hu2021lora}, its proposed setup disobeys both guidelines. The original LoRA updates the whole weight matrix, which is not head-specific. 
To explain this, consider the weight matrix $\mtx{W}_q$ as an example
\footnote{The case of $\mtx{W}_v$ is similar to $\mtx{W}_q$, while for $\mtx{W}_o$, the role of $\mtx{A}$ and $\mtx{B}$ would be exchanged.
For simplicity, we will always assume we are modifying $\mtx{W}_q$ and discuss how to make $\mtx{B}$ head-specific throughout the paper.}. 
In each training step, LoRA updates
$\mtx{W}_q$ with a low-rank matrix $\mtx{BA}$ of the same size as $\mtx{W}_q$.
However, if we denote the $r$-by-$N_h p$ matrix $\mtx{A}$ as $(\mtx{A}^{(1)}, \mtx{A}^{(2)}, \dots, \mtx{A}^{(N_h)})$, 
$\mtx{A}^{(h)} \in \mb R^{r \times p}, \forall i \in [N_h]$,
the modification to the weight matrix $\mtx{W}_q^{(h)}$ for each head would be $\mtx{B} \mtx{A}^{(h)}$'s, $\forall i \in [N_h]$.
We observe that the updates for all the heads share the same column space spanned by $\mtx{B}$.
In the extreme case of rank-1 $\mtx{B}$ (for example), 
the updates for each column in the weight matrix will be in the same direction,
which is not ideal for adapting all the heads. 
Further, as suggested by \citet{hu2021lora}, LoRA evenly assigns the parameter budgets to the weight matrices for $\mtx{Q}$ and $\mtx{V}$,
which deviates from the second guideline. 

\begin{figure}[t]
    \centering
    \includegraphics[width=\linewidth]{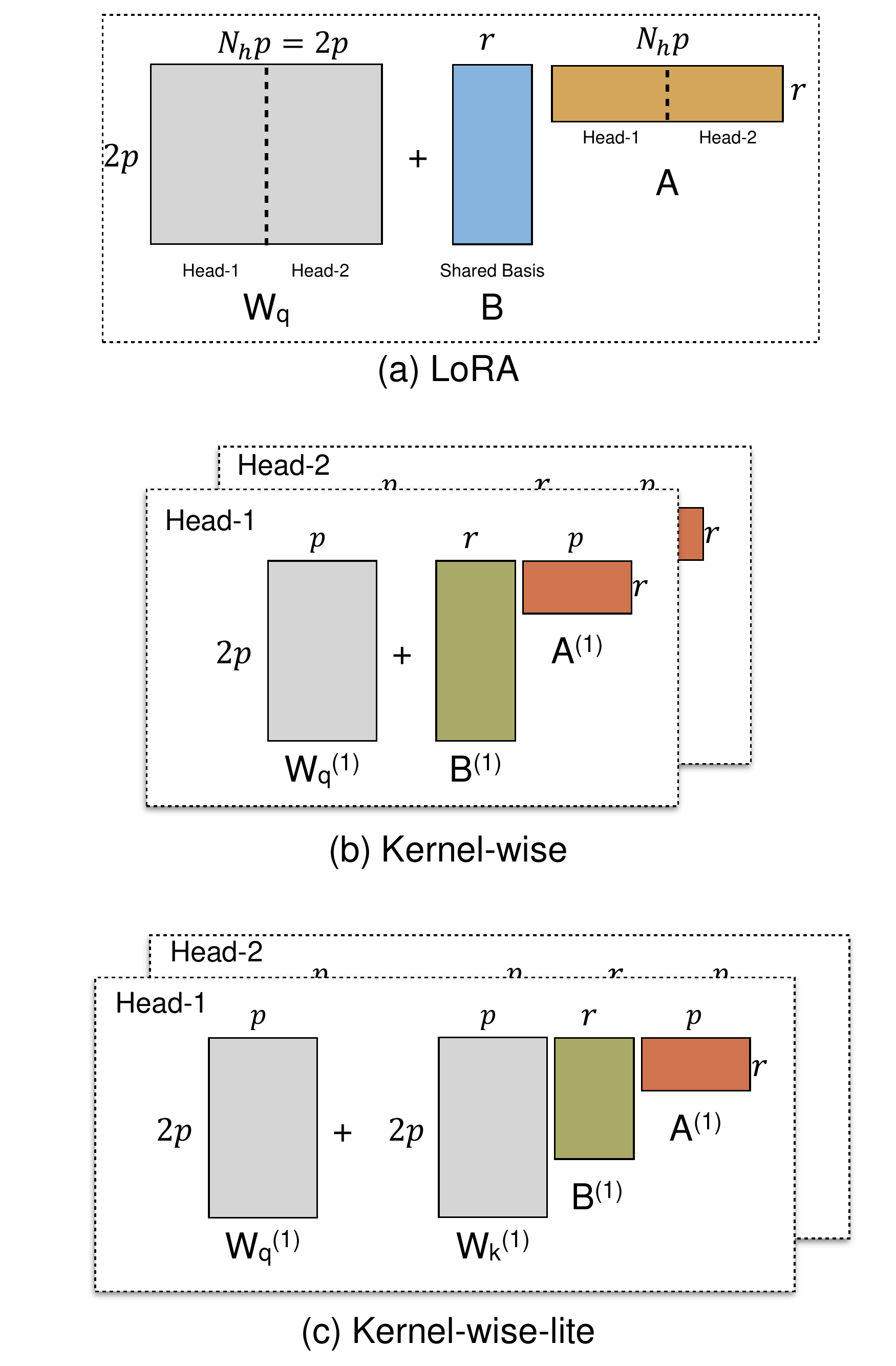}
    \caption{\footnotesize (a) Original LoRa with two heads. (b) \loraspecific~variant of LoRa that satisfies Guideline-1 (c) lightweight version of (b) that uses $\mtx{W}_{k}^{(h)}$ to reduce the trainable parameters in $\mtx{B}^{(h)}$. Note that colored matrices are tunable, whereas gray-scale matrices are fixed (Best viewed in color).}
    \label{fig:LoRA}
\end{figure}

\subsection{Kernel-wise Adaptation}
\label{subsec:method}
We choose LoRA as our primary base to develop our method because of its flexibility in assigning parameters to different weight matrices---both in empirical kernel matrix and coefficient components.

\subsubsection{Guideline-1: Head-specific Adaptation}
\label{sec:guide1}

To incorporate Guideline-1, we extend the framework of LoRA and propose \textbf{\loraspecific}
\footnote{\loraspecific~is a concrete scheme to adjust a specific weight matrix. 
It will have multiple variants modifying different weight matrices.}
that satisfies the first guideline. 

Here, we allow the low-rank weight matrix updates for each head to have customized column spaces, by training distinct $\mtx{B}^{(h)} \in \mb R^{N_h p \times r}$ for head-$h$, $\forall h \in [N_h]$. 
In this case, the weight matrix $\mtx W^{(h)}$ in head-$h$ would be updated by 
\begin{align*}
\mtx W^{(h)} \leftarrow \mtx W^{(h)} + \mtx{B}^{(h)} \mtx{A}^{(h)},
\end{align*}
and is expected to be more expressive due to the non-shared column spaces (see \Cref{fig:LoRA}(b)).

On the downside, this design suffers from inflexibility with a small parameter budget. 
For all the $\mtx{B}^{(h)}$'s, to provide rank-$r$ updates in each head, the new adaptation takes around $N_h^2 p r$ parameters. 
However, if, for instance, only $4 N_h p$ parameters are assigned to modulate a weight matrix, 
we can still implement the original LoRA by using rank-$2$ $\mtx{A}, \mtx{B}$,
while the construction of \loraspecific~would be prohibited since even rank-$1$ updates in each head will require more parameters than the budget.

To resolve the issue, we provide a lightweight alternative to the head-specific adaptation above, which we call \textbf{\loraspecificlite}. In this version, we propose to use the frozen $\mtx{W}_k^{(h)} \in \mb R^{N_h p \times p}$ as the head-specific basis for head-$h$ ($\mtx{W}_k^{(h)}$ is the $h$-th block in the weight matrix $\mtx{W}_k$, c.f.~\Cref{sec:transformer}).
Therefore any target weight matrix $\mtx{W}^{(h)}$ would be updated by 
\begin{align*}
\mtx W^{(h)} \leftarrow \mtx W^{(h)} + \mtx{W}_k^{(h)} \mtx{B}_k^{(h)} \mtx{A}^{(h)},
\end{align*}
where $\mtx{B}_k^{(h)}$ is a $p$-by-$r$ matrix (see \Cref{fig:LoRA}(c)).

Utilizing this $\mtx{W}_{k}^{(h)}$ with a smaller $\mtx{B}_k^{(h)}$ allows the adaptation to be head-specific while containing the same number of trainable parameters as the original LoRA, which is $2 N_h p r$. 
This comes at the cost of restricting the basis spaces of updates for each head from the unconstrained $\mtx{B}^{(h)}$ to $\mtx{W}_k^{(h)} \mtx{B}_k^{(h)}$, where $\mtx{W}_k^{(h)}$ is fixed.

\textit{But why should we choose $\mtx{W}_k$ for the lightweight updates?} The motivation comes from results in kernel learning that encourage adapting of the \textit{coefficient} part using the basis spaces spanned by \textit{key} matrices. 
In a kernel estimator, the coefficient $\mtx{C}$ is independent of the query sequence $\mtx{Q}$ as it is trained solely by the supporting points $\mtx{K}$---asymptotically, $c_j$, the $j$-th row in $\mtx{C}$, is only decided by $k_j, j \in [N]$~\citep{yang2017frequentist}.
Concretely, given the loss function and the kernel function, $c_j$ is in general influenced by $k_j$ and $\mtx{K}_{-j}$ (all the points except $k_j$), 
while when $N \to \infty$, $\mtx{K}_{-j}$ can be fully specified by a fixed distribution.
This implies that in attention, compared to \textit{queries}, \textit{keys} are more related to \textit{values}, 
and following which, we turn to $\mtx{W}_k$ to form the basis for the low-rank updates for the conceptual motivation.

\paragraph{Combining LoRA with \loraspecific.}

Our proposed adaptation can make fine-grained adjustments to each attention head and thus improve the representations. 
However, increased representation power might at times trade-off with lower-rank updates.
As such, we propose our main variant---\textbf{\loramix}, 
to combine the original LoRA and \loraspecific, to attain the best of both worlds---larger basis (therefore higher rank updates) and specific adaptation to each head.
Its update expression is as follows,
\begin{align*}
\mtx W^{(h)} \leftarrow \mtx W^{(h)} + (\mtx{B}_{LoRA}, \mtx{B}^{(h)}) 
\begin{pmatrix}
     \mtx{A}_{LoRA}^{(h)} \\
    \mtx{A}^{(h)}
\end{pmatrix},
\end{align*}
where $\mtx{B}_{LoRA}$ is shared among all the heads, 
while $\mtx{B}^{(h)}$'s are head-specific.
We remark \loramix~also has a lightweight alternative, \textbf{\loramixlite},
which is the combination of the original LoRA and \loraspecificlite.
We compare different variants through experiments in~\Cref{sec:results}.

\subsubsection{Guideline-2: Making Coefficients more Expressive}

To incorporate Guideline-2, we propose to make coefficients more expressive by allowing the modification of both $\mtx{W}_v$ and $\mtx{W}_o$ for the coefficient part, compared to only updating $\mtx{W}_v$ in the original LoRA.
We achieve this by re-writing the attention sub-layer under the kernel estimator perspective,
which extends the scope of attention by including $\mtx{W}_o$ in its head-specific computation as well.
If we represent $\mtx{W}_o$ as,
\begin{equation*}
\begin{pmatrix}
    &\mtx{W}_o^{(1)} \\
    &\vdots \\
    &\mtx{W}_o^{(N_h)}
\end{pmatrix},
\end{equation*}
where, the $i$-th block $\mtx{W}_o^{(h)} \in \mb R^{p \times N_h p}$ corresponds to head-$h$ (i.e., $\mtx{L}^{(h)} \mtx{V}^{(h)}$) in the attention output matrix, we can re-write the attention sub-layer as, 
\begin{align}
\sum_{h=1}^{N_h} \mtx{L}^{(h)} \mtx{V}^{(h)} \mtx{W}_o^{(h)}, \label{eqn:rewritten_attn}
\end{align}
and propose to take each summand as the complete form of a head (kernel estimator).
We thus extend the coefficient part from the value matrices to the matrix products $\mtx{V}^{(h)} \mtx{W}_o^{(h)}$'s, 
which naturally results in $N$-by-$N_h p$ coefficients (with rank-$p$).

\subsection{Final Model}
\label{sec:final-model}

Combining the two pieces together, we report the concrete adaptation scheme under two settings:

\begin{itemize}[leftmargin=*]
\item With very limited parameter budgets (less than 0.2\% of the total PLM parameters), 
similar to LoRA, we modify $\mtx{W}_q$ and $\mtx{W}_v$ with equal parameter budgets using \loramixlite (qv). 
(The suffix (qv) means the method will adjust $\mtx{W}_q$ and $\mtx{W}_v$.)
In this case, we omit Guideline-2 and only incorporate Guideline-1 to apply head-specific updates.
\item With intermediate parameter budgets (around 1.6\% of the total parameters in the PLM), 
we suggest using \loramix (qvo), which instead modifies $\mtx{W}_q$, $\mtx{W}_v$, and $\mtx{W}_o$, assigning more budgets to the coefficient part (Guideline-2). 
Given the increased parameter budgets, we allow \loramix~scheme for $\mtx{W}_q$ and $\mtx{W}_o$,
while continue to utilize \loramixlite~scheme for $\mtx{W}_v$. 
\loramix (qvo) incorporates both the guidelines.
\end{itemize}

\section{Experiments}
While parameter-efficient tuning methods have been extensively studied for NLU tasks, their applicability towards NLG tasks is not well-known. This section performs empirical experiments of our proposed methods on three NLG tasks.
To show the consistent effectiveness of our methods, we provide results on two NLU tasks as well.
\footnote{The code for our algorithms is publicly available in this \href{https://github.com/ychen-stat-ml/kernel-adapters}{repository}. 
The release info is available in this \href{https://www.amazon.science/publications/empower-parameter-efficient-transfer-learning-by-recognizing-the-kernel-structure-in-attention }{page}.}

\begin{table*}[h!]
    \centering
    \resizebox{\textwidth}{!}{
    \begin{tabular}{|lcc|ccccccccc|cc|c|}
    
        \hline
        &&& \multicolumn{9}{c}{\textbf{WebNLG}} & \multicolumn{2}{|c|}{\textbf{CoQA}} & \multicolumn{1}{c|}{\textbf{SUM}} \\
        &\multicolumn{2}{c|}{Parameters to}& \multicolumn{3}{c}{BLEU $\uparrow$}  & \multicolumn{3}{c}{MET $\uparrow$} & \multicolumn{3}{c|}{TER $\downarrow$}  & EM & F1 & \multicolumn{1}{c|}{ROUGE-2} \\
        & train & store  & S & U & A & S & U & A & S & U & A & \multicolumn{2}{c|}{ } &  \\
        \hline
        \multicolumn{15}{c}{\textit{Full Budget $= 100\%$}}
        \\ \hline
        Fine-tuning       & 100.00\% & 100.00\% & 59.8          & 28.7          & 46.1          & 0.43          & 0.29          & 0.36          & 0.38          & 0.68          & 0.51          & 59.0          & 67.4          & 15.72          \\
        \hline
        \multicolumn{15}{c}{\textit{Tiny Budget $< 0.1\%$}}
        \\ \hline
        Adapter-4         & 0.08\%   & 0.08\%   & 51.7          & 35.6          & 44.4          & 0.38          & \textbf{0.32} & \textbf{0.35} & 0.44          & 0.58          & 0.51          & 49.9          & 58.8          & 14.18          \\
        Compacter         & 0.08\%   & 0.08\%   & \textbf{53.3} & 35.0          & \textbf{45.0} & \textbf{0.39} & 0.31          & \textbf{0.35} & \textbf{0.43} & 0.58          & 0.50          & 51.4          & 59.9          & 14.23          \\
        Bitfit            & 0.08\%   & 0.08\%   & 49.0          & 34.9          & 42.6          & 0.37          & \textbf{0.32} & 0.34          & 0.45          & 0.57          & 0.51          & 51.2          & 59.8          & 13.66          \\
        \loramixlite (qv)     & 0.07\%   & 0.07\%   & 51.0          & \textbf{36.7} & 44.5          & 0.38          & \textbf{0.32} & \textbf{0.35} & \textbf{0.43} & \textbf{0.54} & \textbf{0.48} & \textbf{53.1} & \textbf{61.6} & \textbf{14.27} \\
        \hline
        \multicolumn{15}{c}{\textit{Small Budget $< 0.2\%$}}
        \\ \hline
        Adapter-8         & 0.14\%   & 0.14\%   & \textbf{54.8} & 36.4          & \textbf{46.5} & \textbf{0.40} & \textbf{0.33} & \textbf{0.36} & 0.42          & 0.58          & 0.49          & 51.9          & 60.7          & 14.42          \\
        Prefix-tuning-8   & 7.92\%   & 0.12\%   & 49.2          & 35.6          & 43.0          & 0.37          & 0.31          & 0.34          & 0.45          & 0.56          & 0.50          & 50.1          & 58.6          & 14.18          \\
        LoRA-4            & 0.13\%   & 0.13\%   & 52.8          & 37.1          & 45.8          & 0.39          & \textbf{0.33} & \textbf{0.36} & 0.42          & 0.55          & \textbf{0.48} & \textbf{56.1}          & \textbf{64.7}          & 14.42          \\
        \loramixlite (qv)    & 0.13\%   & 0.13\%   & 53.8          & \textbf{37.2} & 46.3          & 0.39          & \textbf{0.33} & \textbf{0.36} & \textbf{0.41} & \textbf{0.54} & \textbf{0.48} & 55.4 & 63.9 & \textbf{14.52} \\ 
        \hline
        \multicolumn{15}{c}{\textit{Intermediate Budget $< 2\%$}}
        \\ \hline
        Adapter-108       & 1.62\%   & 1.62\%   & 59.5          & 34.1          & 48.2          & 0.42          & 0.32          & \textbf{0.38} & 0.38          & 0.61          & 0.49          & 57.7          & 66.4          & 15.22          \\
        Prefix-tuning-108 & 7.98\%   & 1.60\%   & 56.1          & \textbf{37.2} & 47.6          & 0.40          & \textbf{0.33} & 0.37          & 0.40          & \textbf{0.55} & 0.47          & 51.8          & 60.3          & 14.81          \\
        LoRA-54           & 1.61\%   & 1.61\%   & 54.8          & 36.9          & 46.7          & 0.40          & \textbf{0.33} & 0.37          & 0.41          & \textbf{0.55} & 0.47          & 57.2          & 65.7          & 15.29          \\
        \loramix (qvo)        & 1.61\%   & 1.61\%   & \textbf{59.8}* & 36.7          & \textbf{49.3}* & \textbf{0.43}* & \textbf{0.33} & \textbf{0.38} & \textbf{0.37}* & 0.57          & \textbf{0.46}* & \textbf{59.9}*  & \textbf{68.4}*   & \textbf{15.34}* \\
        \hline
        \end{tabular}
    }
        
    \caption[Caption of table]{\label{tab:result} Performance (\%) on NLG tasks \textsuperscript{a}. The methods are divided into 4 groups based on the number of parameters to store, and the methods in the same group have similar sizes. We \textbf{boldface} the best score in each group for different metrics. 
    In the group of intermediate budget, we further conduct the significance tests between Kernel-wise and the best baseline for each metric (* means the test p-value $<0.05$).
    }
    \small\textsuperscript{a} We follow the notations in prefix-tuning \citep{Li2021PrefixTuningOC} that S, U, A represent SEEN, UNSEEN, and ALL respectively; 
    \underline{S}EEN categories are used in training; 
    \underline{U}NSEEN categories only appear in the test set; 
    and \underline{A}LL consists of all the categories.
\end{table*}

\subsection{Experimental Setup}
\label{sec:task}

We mainly evaluate the performance of our methods on NLG tasks, in which there is still a gap between fine-tuning and most parameter-efficient adaptation techniques~\citep{he2021towards}.
Specifically, we conduct our experiments on the following datasets:
WebNLG-challenge~\citep{gardent2017webnlg} for table-to-text tasks,
CoQA~\citep{reddy2019coqa} for conversational question answering,
and CNN/Daily-Mail (CNN/DM)~\citep{hermann2015teaching} for summarization (SUM).
CNN/DM has only one domain, while CoQA has five domains
\footnote{We use the official dev set as the test set and randomly select 500 examples from the train set as the new dev set.},
and WebNLG has 14 domains. Descriptions of the datasets and evaluation metrics are provided in \Cref{sec:appendix_dataset}.

Our experiments mainly follow the setting used by~\citet{lin2020exploring},
which takes $\text{GPT-2}_{\text{SMALL}}$ (124M parameters)~\citep{wolf2019huggingface} as the backbone for all the NLG tasks.
We choose the smaller model size as compared to the larger models used in other related studies it is generally difficult for smaller models to attain the same performance as full-model fine-tuning \citep{Lester2021ThePO, DBLP:journals/corr/abs-2110-07602}. This creates a challenging testbed to evaluate our proposed approaches.

In addition to the NLG experiments, 
we also study two NLU tasks: MNLI \citep{N18-1101} and SST2 \citep{socher2013recursive}, 
to show the performance of our methods on encoder-only transformers. 
The Multi-Genre Natural Language Inference (MNLI) Corpus (sentence pairs of hypotheses and premises with entailment annotations) will be given, and the task is to predict whether the premise entails, contradicts, or is neutral to the hypothesis;
the Stanford Sentiment Treebank (SST2) is composed of movie reviews and corresponding human-annotated sentiment and specifies a task to predict the sentiment of a review sentence (positive/negative).
We implement the backbone under the setting used by \citet{he2021towards} and use $\text{RoBERTa}_{\text{BASE}}$ (125M parameters)~\citep{liu2019roberta} for both MNLI and SST2.

\subsection{Baselines}
\label{sec:candidate}

We compare our method with several other representative methods:
fine-tuning \citep{howard2018universal}, adapters \citep{houlsby2019parameter} used by \citet{lin2020exploring}, Compacter \citep{mahabadi2021compacter},
Bitfit~\citep{ben2021bitfit}, prefix-tuning \citep{Li2021PrefixTuningOC}, LoRA~\citep{hu2021lora}, and MAM-adapter~\citep{he2021towards}.
In Table~\ref{tab:result} we use a postfix of adapters / LoRA / prefix-tuning to indicate their bottleneck size / rank of updates / prefix length, respectively. For instance, Adapter-4 means that the bottleneck size of the two-layer MLP in the inserted adapter is 4. 

For some adaptation techniques, the number of parameters to train is not flexible to tune. For instance, Bitfit proposes to tune all the bias terms within the PLMs, and, as a result, the maximum parameters to tune are limited;
as for Compacter, the weight matrices in the adapter modules are constructed through the Kronecker product, and the parameter complexity is $\m O(\frac{L}{n} + n^3)$ \citep{mahabadi2021compacter}, 
where $n$ is the size of a square matrix used in the Kronecker product and $L$ is the number of layers.
We choose a particular setting to make the number of trainable parameters in Compacter close to Bitfit and a tiny size adapter.
The parameter size of Compacter is not further increased since a larger $n$ would significantly retard the training.

For prefix-tuning, \citet{Li2021PrefixTuningOC} suggest to utilize a re-parametrization trick to mitigate its initialization issue,
and therefore, the number of parameters to train will be much larger than the actual number of parameters to store,
while these two numbers are the same for all other methods.
In deciding the model size, we manage to make the number of parameters to \textbf{store} in prefix-tuning roughly the same as its adapter counterpart by adjusting the prefix length.

\section{Results}
\label{sec:results}

\subsection{Main Results}

Table~\ref{tab:result} compares our proposed methods against other baselines on the aforementioned generation tasks. 
The performance of our proposed \loramix~method on text classification is reported in Table~\ref{tab:nlu}.
We summarize our observations as follows.

\paragraph{Tasks with long input.} 
As shown in Table~\ref{tab:result}, 
all previous parameter-efficient methods fail to attain comparable performance to fine-tuning on CoQA and CNN/DM tasks which have a longer input than the table-to-text generation task WebNLG. This indicates that current parameter-efficient methods still fall behind fine-tuning in those more challenging generation tasks with longer sequences. 
However, \loramix, encloses this gap and even outperforms fine-tuning in some NLG tasks, e.g. the CoQA task, with solely 1.61\% tunable parameters of $\text{GPT-2}_{\text{SMALL}}$.

\paragraph{Impact of parameter sizes.}
The results in Table~\ref{tab:result} show that overall, as the tunable parameter size increases, the performance of various parameter-efficient methods also increases, getting closer to fully fine-tuning. 
This indicates that a large enough parameter budget is still a prerequisite for the excellent performance of parameter-efficient adaptation methods. This finding can help us explain why the performance of Compacter can be better than Adapter-4 over all the tasks when the parameter budget is tiny (\textit{Tiny Budget $<0.1\%$} in the second group in Table \ref{tab:result}), considering that the Compacter can construct a larger MLP than the adapter with the same parameter budget due to the usage of Kronecker product.
Inspired by this finding, we can also clearly see the limitation of Bitfit and Compacter as their parameter budgets are constrained to be small and cannot be elevated of free will. 
As for the original LoRA, the impact of parameter sizes is somewhat tricky---LoRA-4 shows competitive performance while LoRA-54 is not improved as greatly as other methods on WebNLG and CoQA.
A similar phenomenon on different datasets is also observed by \citet{hu2021lora, he2021towards}.

\paragraph{Performance of our proposed \loramix (qvo).}
Our proposed \loramix (qvo) can generally improve the performance on all three NLG datasets.
On WebNLG, \loramix (qvo) provides a 1.1 increase in BLEU score compared to Adapter-108 and a 2.6\% increase compared to LoRA-54;
on CoQA, our method is even more greatly better than LoRA-54, obtaining 2.7 exact-match and F1 improvement, 
and even outperforms fine-tuning by around 1\% in both of the metrics;
On CNN/DM, all the parameter-efficient methods have close performance, while through a test, we show our method Kernel-mix has a significantly higher Rouge-2 score than the best baseline, LoRA-54.
Overall, Table~\ref{tab:result} demonstrates that kernel-mix adaptation better exploits the attention structure in PLMs and improves the overall generation quality under all three kinds of parameter budgets.

\begin{table}[h!]
    \centering
    \begin{tabular}{|l|c|c|}
\hline 
Method (\# params) & MNLI & SST2 \\
\hline 
Fine-tuning (100\%) & $87.6_{\pm .4}$ & $94.6_{\pm .4}$ \\
\hline 
Bitfit $(0.1 \%)$ & $84.7$ & $93.7$ \\
Prefix-tuning $(0.5 \%)$ & $86.3_{\pm .4}$ & $94.0_{\pm .1}$ \\
LoRA $(0.5 \%)$ & $87.2_{\pm .4}$ & $94.2_{\pm .2}$ \\
Adapter $(0.5 \%)$ & $87.2_{\pm .2}$ & $94.2_{\pm .1}$ \\
MAM-Adapter $(0.5 \%)$ & $87.4_{\pm .3}$ & $94.2_{\pm .3}$ \\
\hline
\loramix (qvo) $(0.5 \%)$ & $87.4_{\pm .2}$ & $94.3_{\pm .3}$ \\
\hline
\end{tabular}

\caption{\label{tab:nlu} 
Accuracy on the dev set of MNLI and SST2. Bitfit numbers are copied from \citet{ben2021bitfit}, and all the other results (except for \loramix) are from \citet[Table~2]{he2021towards}.
}
\end{table}

\paragraph{Performance on NLU tasks.}
Table~\ref{tab:nlu} shows the performance of \loramix (qvo) when it is extended to encoder-only transformers.
For a fair comparison, we specify a new parameter budget (0.5\%) for \loramix (qvo), different from the previous settings in Table~\ref{tab:result}.
With the new budget, \loramix (qvo) attains close accuracy to the other parameter-efficient methods on both MNLI and SST2.
We remark that the parameter budget used here is slightly tight for \loramix (qvo), as the ranks assigned for head-wise adaptation ($1$ for $W_q$ and $2$ for $W_v, W_o$) are limited (see Table~\ref{tab:qvo}).

\begin{table*}[h!]
    \centering
    \resizebox{\textwidth}{!}{
    \begin{tabular}{|lcc|ccccccccc|cc|c|}
\hline
&&& \multicolumn{9}{c}{\textbf{WebNLG}} & \multicolumn{2}{|c|}{\textbf{CoQA}} & \multicolumn{1}{c|}{\textbf{SUM}} \\
&\multicolumn{2}{c|}{Parameters to}& \multicolumn{3}{c}{BLEU $\uparrow$}  & \multicolumn{3}{c}{MET $\uparrow$} & \multicolumn{3}{c|}{TER $\downarrow$}  & EM & F1 & \multicolumn{1}{c|}{ROUGE-2} \\
& train & store  & S & U & A & S & U & A & S & U & A & \multicolumn{2}{c|}{ } &  \\
\hline
\multicolumn{15}{c}{\textit{Budget $< 0.2\%$}}
\\ \hline
LoRA-4                & 0.13\% & 0.13\% & 52.8   & 37.1   & 45.8   & 0.39  & 0.33  & 0.36 & 0.42  & 0.55  & 0.48  & 56.1 & 64.7 & 14.42    \\
\loraspecificlite (qv) & 0.13\% & 0.13\% & \textbf{55.1}   & 36.9   & \textbf{46.9}   & \textbf{0.40}  & 0.33  & \textbf{0.37} & \textbf{0.40}   & 0.55 & \textbf{0.47}  & 55.0 & 63.6 & \textbf{14.48}    \\
\hline
\multicolumn{15}{c}{\textit{Budget $< 2\%$}} \\
\hline
Kernel-wise(mq)    & 1.56\% & 1.56\% & 58.9 & 36.7 & 48.9 & 0.42 & 0.33 & 0.38 & 0.37 & 0.57 & 0.46 & 57.4 & 66.3 & 15.22 \\
Kernel-wise(mv)    & 1.56\% & 1.56\% & 59.4 & 37.2 & 49.3 & 0.42 & 0.33 & 0.38 & 0.37 & 0.57 & 0.46 & 58.0 & 67.0 & 15.24 \\
Kernel-wise(qvo)   & 1.61\% & 1.61\% & 59.5 & 36.1 & 49.0 & 0.43 & 0.33 & 0.38 & 0.37 & 0.58 & 0.47 & 59.1 & 68.0 & 15.28 \\
Kernel-mix(qvo)    & 1.61\% & 1.61\% & 59.8 & 36.7 & 49.3 & 0.43 & 0.33 & 0.38 & 0.37 & 0.57 & 0.46 & 59.9 & 68.4 & 15.34 \\
\hline
\end{tabular}
}
\caption{\label{tab:ablation} 
Performance on new variants of \loraspecific~compared to LoRA-4 and our proposed \loramix (qvo) (both copied from Table~\ref{tab:result}). 
The methods with similar budgets of tunable parameters are grouped.
The exact settings of the methods to compare are illustrated in \Cref{sec:ablation} and Appendix~\ref{sec:method_setting}.
}
\end{table*}

\subsection{Ablation Studies}
\label{sec:ablation}

Besides the main results in Table~\ref{tab:result}, we also perform ablation studies to verify the effectiveness of our propositions.
We additionally implement four variants of \loraspecific~to help ablate the effects of our proposed guidelines.
Among the new variants, \loraspecificlite (qv), \loraspecific (mq), and \loraspecific (mv) only adjust $\mtx{W}_q$, $\mtx{W}_v$;
\loraspecificlite (qv) takes the strategy in LoRA-4 to evenly assign parameters to $\mtx{W}_q$ and $\mtx{W}_v$;
\loraspecific (mq) leaves more budget to $\mtx{W}_q$ than $\mtx{W}_v$ with a ratio of 3:1, 
while \loraspecific (mv) is set up in the reversed way.
In contrast, \loraspecific (qvo) simultaneously adjusts $\mtx{W}_q$, $\mtx{W}_v$, and $\mtx{W}_o$ (with a budget ratio of 5:1:10).
\di{with an equal budget?}
The experimental results are summarized in Table~\ref{tab:ablation}. 
The settings of the variants designed for ablation are described in Appendix~\ref{sec:method_setting}.

\textbf{Head-specific adaptation} (Guideline-1).
For ``LoRA-4'', the setting recommended by \citet{hu2021lora}, we compare it with its head-specific counterpart---\loraspecificlite (qv).
In almost all the tasks, \loraspecificlite (qv) can attain better performance with the same number of parameters.

\textbf{More parameters for the coefficient part} (Guideline-2).
\citet{hu2021lora} (Section~7.1) indeed have already done some preliminary exploration to find 
the relatively more important weight matrices in transformers.
Their experimental results (copied as Table~\ref{tab:lora_copy} in Appendix~\ref{sec:loraTable}) clearly show that ``putting all the parameters in $\Delta \mtx{W}_q$ or $\Delta \mtx{W}_k$ results in significantly lower performance''.
In this work, we additionally show that by simply moving some trainable parameters from $\mtx Q$, the empirical kernel matrix part, 
to $\mtx V$, the coefficient part, 
\loraspecific (mv) can improve the performance upon \loraspecific (mq) as well.

\textbf{Extending the scope of attention}.
To show the benefits of both adjusting $\mtx{W}_v$ and $\mtx{W}_o$, 
we compare the new variant \loraspecific (mv) against \loraspecific (qvo).
They both assign more budgets to the coefficient part;
\loraspecific (qvo) would update both $\mtx{W}_q, \mtx{W}_v$ and $\mtx{W}_o$, 
while \loraspecific (mv) only adjusts $\mtx{W}_q, \mtx{W}_v$.
We can observe that \loraspecific~has better performance in most tasks.

\textbf{Combining the shared and the head-specific basis}.
Lastly, we find that \loramix (qvo) (our proposed method) outperforms \loraspecific (qvo), which justifies combining the two types of basis, as opposed to pure head-specific adaptation.

\section{Conclusion and Future Work}

In this work, we revisit the connection between the attention module and kernel estimators, and accordingly propose kernel-wise adaptation,
which adopts the guidelines from kernel learning to strengthen the low-rank adaptation (LoRA).
We verify that with the same parameter budgets, our proposed adaptation techniques can have better performance on three generation tasks than the existing parameter-efficient methods, including adapters, prefix-tuning, and LoRA, and attain close accuracy on two classification tasks as well.

One possible extension of our work is combining our proposed method with other adapters in feed-forward sub-layers. 
In MAM-adapter, \citet{he2021towards} suggest applying prefix-tuning to adapt the parameters in self-attention sub-layers and assigning budgets to feed-forward sublayers as well;
it can be beneficial to replace prefix-tuning with \loramix~for adaptation in the attention part.

Another direction of future research is the extension of our method to the feed-forward sub-layers, which are interpreted as key-value memories in recent work and behave like attention blocks~\cite{geva2021transformer}. It will be interesting to study if kernel-specific guidelines could help design better adapters employed in the feed-forward layers.

\section*{Acknowledgements}

We appreciate all the valuable feedback from the anonymous reviewers.

\bibliography{custom}
\bibliographystyle{acl_natbib}

\clearpage
\appendix

\renewcommand\thetable{\Alph{section}.\arabic{table}}
\renewcommand\thefigure{\Alph{section}.\arabic{figure}}

\section{Dataset Details}
\label{sec:appendix_dataset}

\begin{itemize}[leftmargin=*]
\item The \textbf{WebNLG} dataset consists of mapping sets of RDF triples to text.
The training data are Data/Text pairs where the data is a set of (subject, property, object) triples.
There are 9 categories extracted from DBpedia in the train and the development (dev) set,
while the test set contains 5 more unseen categories,
which can be used to evaluate the generalization of the adaptation methods.
We adopt the official evaluation script and reports BLEU \citep{papineni2002bleu}, METEOR, \citep{lavie2007meteor} and TER \citep{snover2006study}
\footnote{For TER, the lower the metric is, the better the performance is.}.
\item \textbf{CoQA} is a large-scale conversational question answering dataset. 
It contains over 127K questions with answers collected from more than 8K conversations.
The problem involves generating answers to the questions based on related conversation histories and documents.
We follow the official evaluation script and use the macro-average F1 score of word overlap as the main evaluation metric~\citep{reddy2019coqa}.
\item \textbf{CNN/DM} is a benchmark for text summarization, involving more than 300K news articles provided by CNN and the Daily Mail.
We report ROUGE-2 scores \citep{lin2004rouge} as evaluation metrics.

\item \textbf{MNLI} The Multi-Genre Natural Language Inference Corpus~\citep{N18-1101} provides sentence pairs of hypotheses and premises with entailment annotations.
There are $393k$ pairs in the training set, $10k$ in the dev set, and another $10k$ pairs in the test set.
(Only the dev set is used in Table~\ref{tab:nlu}.) 
The premise sentences come from ten different sources, 
and the model performance can be evaluated on both the matched (in-domain) and mismatched (cross-domain) sections.
In Table~\ref{tab:nlu}, we follow the setting used by \citet{hu2021lora} and report mismatched accuracy as the metric.
\item \textbf{SST2} The Stanford Sentiment Treebank \citep{socher2013recursive} is a corpus with fully labeled parse trees. 
In this corpus, $11,855$ single sentences extracted from movie reviews were parsed with the Stanford parser \citep{klein2003accurate}, generating $215,154$ unique phrases from those parse trees (3 human judges annotate each phrase). 
\citet{wang2019glue} incorporated the task into the GLUE benchmark, with $67k$ sentences in the training set, $0.9k$ in the dev set, and $1.8k$ instances in the dev set. 
(Only the dev set is used in Table~\ref{tab:nlu}.)
The metric is the accuracy of the decision whether the sentiment of a review sentence is positive or negative.
\end{itemize}

\section{Training Details}
\label{sec:appendix_training}

\begin{table*}[h!]
\centering
\resizebox{\textwidth}{!}{
\begin{tabular}{|l|c|c|}
\hline
\textbf{Datasets}        & \textbf{Special tokens}       & \textbf{\# of trainable parameters for task embedding} \\
\hline
WebNLG          &
\begin{tabular}{@{}c@{}}
\texttt{<bos\_webnlg>}, \texttt{<eos\_webnlg>}, \texttt{<subject>}, \\
\texttt{<property>}, \texttt{<object>}, \texttt{<target\_webnlg>}
\end{tabular}
                                        & $6 * 768 = 4608$          \\ \hline
CoQA            &
\begin{tabular}{@{}c@{}}
\texttt{<bos\_qa>}, \texttt{<eos\_qa>}, \texttt{<question>}, \\
\texttt{<answer>}, \texttt{<document>}
\end{tabular}
                                        & $5 * 768 = 3840$    \\ \hline
CNN/DM  &
\begin{tabular}{@{}c@{}}
\texttt{<bos\_sm>}, \texttt{<eos\_sm>}, \texttt{<source\_sm>}, \\
\texttt{<target\_sm>}
\end{tabular}
                                        & $4 * 768 = 3072$        \\ \hline
\end{tabular}
}
\caption{\label{tab:tokens} The special tokens used in different tasks and the corresponding size of trainable parameters. 
}
\end{table*}

\subsection{General Training Settings}
\label{sec:task_embedding}

We avoid the sentence-level knowledge distillation trick used in~\cite{lin2020exploring}, as it might interfere in analyzing our hypotheses. However, we retain the usage of ``task embeddings'' as they are required by the original GPT-2 model. These task embeddings act as specialized segment embeddings that indicate the different components of the text input (e.g., the three components of a triple in NLG, questions and answers in CoQA, etc.).
\footnote{The task embedding for the special tokens will also be updated during training, while we do not count them in~\Cref{tab:result}.} 

We state the specific task embedding used in each task.
For CoQA and CNN/DM, we follow the task embedding suggested by \citet{lin2020exploring};
for WebNLG, we similarly set up the special tokens for the different components in the triples.
We conclude the details of the special tokens in each dataset in Table~\ref{tab:tokens}.
Notably, the parameter budget for task embedding is neglectable compared to the size of the aforementioned parameter-efficient adaptation techniques.

\subsection{Hyper-parameters}

We apply an AdamW optimizer and a linear learning rate scheduler with a 500-step warmup duration in training. At generation time, we use a greedy search for all the tasks, the same as \citet{lin2020exploring}.
For the choice of some hyperparameters, we mainly follow the setting used by \citet{lin2020exploring, hu2021lora} and \citet{he2021towards}, including the number of epochs and the argument for weight decay.
Specifically, for WebNLG, CNN/DM, MNLI, and SST2, we train the model for 10 epochs; 
for CoQA, we train the model for 5 epochs.
For the other important hyper-parameters, such as batch size and learning rate, 
we tune the hyper-parameters for different methods according to the loss on the validation set.
For our \loramix~methods in Table~\ref{tab:result}, they share the same batch size and learning rate in each task.
Specifically, for WebNLG, the learning rate we use is $0.00125$, and the batch size is $16$;
for CoQA, the learning rate we use is $0.005$, and the batch size is $8$;
for CNN/DM, the learning rate we use is $0.001$, and the batch size is $16$;
for MNLI, the learning rate we use is $0.0002$, and the batch size is $32$;
for SST2, the learning rate we use is $0.0001$, and the batch size is $16$;

We train each variant for multiple independent runs to account for variability. In particular, for WebNLG, we train models over 5 runs, for CoQA 3 runs, for CNN/DM 2 runs, for MNLI 3 runs, and for SST2 3 runs.\footnote{We reduce the number of runs for larger datasets given computation budget.} 
The reported numbers in Tables~\ref{tab:result} and \ref{tab:nlu} are the mean value averaged over the runs.

\subsection{Implementation and Training Efficiency}
\label{sec:implementation}

All the models in this work are implemented by PyTorch.
For the devices, we perform the distributed training using 8 Tesla V100 16GB GPUs.
On WebNLG, it will take our method around $1$ minute to finish one epoch;
on CoQA, the time cost is around $20$ minute / epoch;
on CNN/DM, the training time per epoch will be $30$ minutes.
For the NLU tasks, the task implementation by \citet{he2021towards} cannot be adapted to distributed training (can only be trained with one graphic card), and thus the training time is longer:
it will take our method around $2$ hours to finish one epoch in MNLI, and $25$ minutes in SST2.

We additionally report there is actually an implementation trick in \loraspecificlite. 
We can simply associate the $h^{th}$ head's \textit{key} matrix $\mtx{K}^{(h)}$ to the computation of any weight matrix, say $\mtx{Q}^{(h)}$, as follows, 
\begin{align*}
    \mtx{Q}^{(h)} &=\mtx{X}\mtx{W}_{q}^{(h)} + \mtx{X}\mtx{W}_{k}^{(h)}\mtx{B}_k^{(h)} \mtx{A}^{(h)}\\
    &=\mtx{X}\mtx{W}_{q}^{(h)} + \mtx{K}^{(h)} \mtx{B}_k^{(h)} \mtx{A}^{(h)}.
\end{align*}
In that case we can reuse the given $\mtx{K}$ to save the computation of the product 
$\mtx{X} (\mtx{W}_{k}^{(h)}\mtx{B}_k^{(h)} \mtx{A}^{(h)})$.

\subsection{Specific settings for each method}
\label{sec:method_setting}

We report the exact setting for the methods that need further explanation in this subsection.
For Compacter, the bottleneck size of the adapter is $192$, and the number of components is $4$, as suggested by \citet{karimi2021compacter};
for the original LoRA and the variants of our proposed methods, we summarize their settings in Table~\ref{tab:qvo}.
In this table, the numbers in columns Q\_wise, V\_wise, and O\_wise are the rank used for \loraspecific;
if the number is followed by ``(lite)", we apply \loraspecificlite~with the listed rank to adjust the corresponding weight matrices.
The numbers in columns Q\_LoRA, V\_LoRA, and O\_LoRA are the rank of the update used as in the original LoRA.
For \loramix~methods, the numbers in columns Q\_wise and Q\_LoRA (for example) will be non-zero.

\begin{table*}[h]
\centering
\resizebox{\textwidth}{!}{
\begin{tabular}{|l|c|cccccc|}
\hline
                     & Budget       & Q\_wise  & Q\_LoRA & V\_wise   & V\_LoRA & O\_wise   & O\_LoRA \\ 
\hline
LoRA-4               & small        & 0        & 4       & 0         & 4       & 0         & 0       \\
LoRA-54              & intermediate & 0        & 4       & 0         & 4       & 0         & 0       \\
Kernel-mix-lite(qv)  & tiny         & 1 (lite) & 1       & 1 (lite)  & 1       & 0         & 0       \\
Kernel-mix-lite(qv)  & small        & 2 (lite) & 2       & 1 (lite)  & 2       & 0         & 0       \\
Kernel-mix(qvo)      & intermediate & 3        & 12      & 8 (lite)  & 8       & 8         & 8       \\ \hline
Kernel-wise-lite(qv) & small        & 4 (lite) & 0       & 4 (lite)  & 0       & 0         & 0       \\
Kernel-wise(mq)      & intermediate & 12       & 0       & 4         & 0       & 0         & 0       \\
Kernel-wise(mv)      & intermediate & 4        & 0       & 12        & 0       & 0         & 0       \\
Kernel-wise(qvo)     & intermediate & 5        & 0       & 10 (lite) & 0       & 10  & 0       \\ \hline
Kernel-mix(qvo)     & $0.5\%$ in Table~\ref{tab:nlu} & 1        & 2       & 2 (lite) & 4       & 2  & 4 \\
\hline
\end{tabular}
}
\caption{\label{tab:qvo} The exact settings for the original LoRA and the variants of our proposed methods. 
}
\end{table*}

\section{Partial Experimental Results Reported in LoRA \citep{hu2021lora}}
\label{sec:loraTable}

\begin{table*}[h!]
\centering
\begin{tabular}{l|cccccccc}
\hline & \multicolumn{8}{c}{ \# of Trainable Parameters $=18 \mathrm{M}$} \\
\hline Weight Type & $W_{q}$ & $W_{k}$ & $W_{v}$ & $W_{o}$ & $W_{q}, W_{k}$ & $W_{q}, W_{v}$ & $W_{q}, W_{k}, W_{v}, W_{o}$ \\
Rank $r$ & 8 & 8 & 8 & 8 & 4 & 4 & 2 \\
\hline WikiSQL $(\pm 0.5 \%)$ & $70.4$ & $70.0$ & $73.0$ & $73.2$ & $71.4$ & $\mathbf{7 3 . 7}$ & $\mathbf{7 3 . 7}$ \\
MultiNLI $(\pm 0.1 \%)$ & $91.0$ & $90.8$ & $91.0$ & $91.3$ & $91.3$ & $91.3$ & $\mathbf{9 1 . 7}$ \\
\hline
\end{tabular}

\caption{\label{tab:lora_copy} Validation accuracy provided by \citet[Table~5]{hu2021lora} on WikiSQL and MultiNLI.}
\end{table*}

For ease of reading, we copy Table~5 from the paper \citep{hu2021lora} as a piece of evidence to show ``putting all the parameters in $\Delta \mtx{W}_q$ or $\Delta \mtx{W}_k$ results in significantly lower performance''.
\end{document}